\documentclass[sigconf]{acmart}

\usepackage[ruled,vlined]{algorithm2e}

\AtBeginDocument{%
  \providecommand\BibTeX{{%
    \normalfont B\kern-0.5em{\scshape i\kern-0.25em b}\kern-0.8em\TeX}}}

\setcopyright{acmcopyright}
\copyrightyear{2020}
\acmYear{2020}

\acmConference[WebSci '20]{WebSci '20: ACM Conference on Web Science}{July 07--10, 2020}{Southampton, UK}

\begin{document}

\title{An Automated Pipeline for Character and Relationship Extraction from Readers' Literary Book Reviews on Goodreads.com}

\author{Shadi Shahsavari$^1$, Ehsan Ebrahimzadeh$^1$, Behnam Shahbazi$^2$, Misagh Falahi$^1$, Pavan Holur$^1$, Roja Bandari$^1$, Timothy R. Tangherlini$^3$, Vwani Roychowdhury$^1$ }

\affiliation{
\textbf{1} Electrical and Computer Engineering, UCLA, Los Angeles, CA, USA
\\
\textbf{2} Computer Science, UCLA, Los Angeles, CA, USA
\\
\textbf{3} Scandinavian Section, UCLA, Los Angeles, CA, USA
}
 
\renewcommand{\shortauthors}{Shahsavari, et al.}

\begin{abstract}
Reader reviews of literary fiction on social media, especially those in persistent, dedicated forums, create and are in turn driven by underlying narrative frameworks. In their comments about a novel, readers generally include only a subset of characters and their relationships, thus offering a limited perspective on that work. Yet in aggregate, these reviews capture an underlying narrative framework comprised of different actants (people, places, things), their roles, and interactions that we label the ``consensus narrative framework''. We represent this framework in the form of an actant-relationship story graph. Extracting this graph is a challenging computational problem, which we pose  as a latent graphical model estimation problem. Posts and reviews are viewed as samples of sub graphs/networks of the hidden narrative framework. Inspired by the qualitative narrative theory of Greimas, we formulate a graphical generative Machine Learning (ML) model where nodes represent actants, and multi-edges and self-loops among nodes capture context-specific relationships. We develop a pipeline of interlocking automated methods to extract key actants and their relationships, and apply it to thousands of reviews and comments posted on Goodreads.com. We manually derive the ground truth narrative framework from SparkNotes, and then use word embedding tools to compare relationships in ground truth networks with our extracted networks. We find that our automated methodology generates highly accurate consensus narrative frameworks: for our four target novels, with approximately 2900 reviews per novel, we report average coverage/recall of important relationships of \textgreater 80\% and an average edge detection rate of \textgreater 89\%. These extracted narrative frameworks can generate insight into how people (or classes of people) read and how they recount what they have read to others.  \footnote{Our code will be posted at https://github.com/Roychowdhury-group/goodreads} 

\end{abstract}


\maketitle

\section{Introduction}
\subsection{Motivation}
Literary fiction attracts large reading audiences both in the United States and internationally. A National Endowment for the Arts survey reveals that, despite a long steady decline in literary reading in the United States, the number of American adults who read at least one work of fiction a year, even after excluding books read for school or work, still hovers around 43\%. Social media, although often criticized for contributing to the decline in literary reading, has also offered opportunities for communities of readers to interact and engage in ongoing conversations, perhaps thereby reducing the otherwise negative impact of social media on reading. Book forums on social media provide readers an opportunity to share their experiences of reading and can, for some works of fiction, engender long running conversations about nuanced aspects of the work in question. These discussions range from explorations of twists and turns in the plot, to simple declarations of admiration for or familiarity with certain actants (characters, places, things).  \cite{2020bourier}\cite{lehnert1980narrative} \textit{Taken individually}, book commentaries and reviews provide a highly individualized perspective on a work of fiction, focusing only on a few actants and their relevance to the narrative. \textit{Taken together}, these comments provide insight into a broader reader consensus of a novel's overarching narrative framework, comprising a majority of the actants and their nuanced relationships.%

\subsection{Objectives and Challenges}
In our work, we assume that we are given thousands of user reviews of a particular  novel from a social cataloging/review website such as Goodreads.com. Given such a corpus, we ask the following questions: (i) Can one \textit{automatically discover all the primary actants} as well as meta-actants (authors, actors and actresses from film adaptations, etc.) that are mentioned across all of the book reviews for a given novel? (ii) Can one also \textit{discover and meaningfully cluster all the  inter-actant relationships} that these reviews include? The results of goals (i) and (ii) provide, when properly thresholded and weighted, a representation of the consensus model of the novel as perceived by those readers who review the book. Inspired by the actantial narrative model of Algirdas Greimas \cite{greimas1973actants}, we  represent these results as an automatically generated narrative network, where nodes are actants and edges are directed multi-edges annotated with the extracted relationships.  (iii) Finally, \textit{given an expert generated ground truth narrative network}, can one \textit{automatically compare that ground truth network with the auto-generated summary narrative framework network} and compute meaningful metrics such as recall and precision?

Solving the above problems is tantamount to 
developing a view of the reviewers' consensus about a target novel, as readers recollect and review the actual cast of actants and their inter-actant relationships.

The more often that an actant or relationship appears in the corpus, the more heavily it is weighted in the network graph. Importantly, the related methodologies presented here can be extended well beyond the realm of literary fiction to derive narrative frameworks undergirding nearly any collection of documents. We focus on literary fiction because of the unusual (for cultural datasets) presence of a ground truth against which to measure the accuracy of our results. 

To construct the actant relationship narrative graph, we start with a dependency tree parsing of the sentences in each review and extract various syntactic structures, such as the Subject (captured as noun argument phrases), Object (also captured as noun argument phrases), actions connecting them (captured as verb phrases), as well as their alliances and social relationships (captured as explicitly connected adjective and appositive phrases; see Table \ref{tab:appos}; see the Methodology section for the tools used and relationship patterns extracted in this paper). \textit{The task of  aggregating these extracted phrases into a single narrative network poses unique computational challenges}.  

First, as these extractions are both varied and extremely noisy, we need to reduce ambiguity across entity mentions. For example, in reviews of \textit{The Hobbit}, Bilbo Baggins is referred to in numerous ways, including ``Bilbo'' (and its misspelling ``Bilbos''), ``The Hobbit'', ``Baggins'' and ``the Burgler'' or ``the Burglar''. We refer to this disambiguation task as the \textit{Entity Mention Grouping} (EMG) problem. Humans solve the EMG problem by using  context: for the different mentions of a character to be the same, they must have the same relationships with other characters. The human ability to disambiguate in this manner has proven difficult to replicate with computational tools. 

Second, the same challenge applies to inter-actant relationships.  For example, the  relationship ``create'' between Dr. Frankenstein and the monster in the novel \textit{Frankenstein}, can be referred to by a cloud of different phrases, including ``made'', ``assembled'', and ``constructed''. To solve this ambiguity, one must computationally recognize that these words are contextually synonymous and identify the group as constituting a single relationship. To make matters more challenging, there are often numerous different relationships between the same actant pair. The decision tree parsing step produces an unordered list of phrases, which then has to be clustered into semantically similar groups, where each group captures one of the distinct relationships. For example, the extracted relationship phrases between Dr. Frankenstein and the monster include 
\{\textit{created, destroying, kill, regretting, constructed, denied, hates, disgusted, made, assemble, blaming, abandon, runs away}\}. These phrases, however, contain sample phrases from at least three distinct relationships: \underline{Create:} [\textit{created, constructed, made, assemble}], 
\underline{Destroy:} [\textit{destroying, kill}], and \underline{Deny}: [\textit{denied, hates, disgusted, blaming, abandon, runs away, regretting}].  We label this problem of reliably clustering relationships as the \textit{Inter-actant Relationship Clustering} (IARC) problem.

Finally, the task of \textit{quantitative evaluation} -- comparison of  the extracted networks with ground truth networks -- shares many of the same challenges as the previous two tasks. One has to \textit{semantically align} any expert-created network with the automatically created one. For example, one should be able to match an expert annotated relationship of ``X $\rightarrow$ Captured $\rightarrow$ Y,'' to an automatically aggregated relationship, such as ``Y $\rightarrow$\{ escaped, rescued\} from$\rightarrow$ X.''
\subsection{Related Work}
Numerous studies have explored book review collections while several other works have attempted to recreate  story plots based on these reviews \cite{ucsd1, ucsd2, 2020bourier}. The sentence-level syntax relationship extraction task has been studied widely in work on Natural Language Processing and Open Information Extraction \cite{schmitz2012open, fader2011identifying, wu2010open, gildea2002automatic, baker1998berkeley, palmer2005proposition} as well as in relation to the discovery of actant-relationship models for corpora as diverse as conspiracy theories and national security documents \cite{mohr2013,samory2018a}. 
There is considerable recent work on word and phrase embedding for encoding semantic similarity. While word embedding methods such as word2vec, fastText and GloVe \cite{fasttext,glove,word2vec} yield vectors that are context invariant, more recent models such as ELMo and BERT \cite{elmo, bert} allow for polysemy (context-dependent embedding). This polysemic feature allows entire phrases to be encoded to both word-level and phrase-level embedding. We use BERT embedding in this paper.

While there is work, such as Clusty \cite{clusty},  which categorizes entities into different categories in a semi-supervised manner, the category examples are fixed. Similarly, works such as ConceptNet \cite{conceptnet} use a fixed set of selected relations to generate their knowledge base. Other recent entity mention grouping work \cite{google_entity_m} seeks to map entity mentions via context vectors produced as an aggregated feature from high-level document metadata and proximal phrases to the mention within the text. Similar work in story graph applications \cite{chargraph1} create co-scene presence character networks predicated on higher-level annotated knowledge, such as joint scene presence and/or duration of dialogue between a pair of characters. Moreover, these works assume perfect reliability in character mentions (thus obviating the need for the critical step of Entity Mention Grouping that is needed for social media reviews), an assumption we cannot make given our data or data from similarly informal domains.

A major challenge in work on reader reviews of novels is that predefined categories for novel characters and for the diverse inter-character relationships do not exist.  In addition, document level features are missing while the proximal text is sparse due to the inherent size of a review (or tweet, comment, opinion, etc.). An unsupervised scheme such as ours for grouping entity mentions into characters and clustering of relationships into  semantically distinct groups, as an approximate imitation of human processes, has not been addressed previously.

\subsection{Outline of the paper}
In Section 2, we describe our data, our selection of the four novels for analysis, and our method for generating ground truth narrative frameworks. In Section 3 we describe our methodology and how we solve the EMG and IACR problems. In Section 4, we provide an overview of the limitations of our pipeline. In section 5, we present our results and evaluation, and in section 6, we discuss the findings. Lastly, in section 7, we suggest potential improvements that can be incorporated into the pipeline in future work.

\section{Resources}
We use reader reviews of four works of fiction from the community forums on Goodreads: \textit{Frankenstein} (1818); \textit{Of Mice and Men} (1937); \textit{The Hobbit} (1937);  and \textit{To Kill a Mockingbird} (1960) \cite{shelley2015frankenstein,steinbeck1937mice,tolkien2012hobbit,lee1960kill}. The works were chosen from the list of the most frequently rated books on the Goodreads site (number of ratings  $>500,000$). For highly rated novels, the number of reviews is also quite high, although significantly lower than the number of ratings. For example, \textit{The Hobbit} has been rated over $2.5$ million times, but has $44,831$ reviews (at the time of our data collection). For each of the novels, we downloaded the maximum allowed three thousand reviews given the Goodreads API limits on review requests. 
\par
The reviews were harvested using a crawler specifically designed for this project. Not all reviews were useful since numerous posts were either spam, posts on different topics, or written in languages other than English. Other reviews were either too short to include meaningful content, or so garbled as to be unintelligible. After filtering the reviews, we were left with a corpus of 8693 usable reviews: \textit{Frankenstein} (2947), \textit{The Hobbit} (2897), \textit{Of Mice and Men} (2956), and \textit{To Kill a Mockingbird} (2893). We discovered two types of phrases in the reviews: (i) Opinion phrases that reflected the readers’ opinions about the book, the author, or the various characters and events. Relationships extracted from these phrases are the dominant ones when aggregated over all readers’ posts, which is not surprising given that these posts are intended to be reviews. (ii) Plot phrases that describe what happened to a subset of the actants, and how they interacted with each other. These phrases contain both the actants and their relationships, and are of primary interest to us.
\begin{table}[]
    \centering
    \begin{tabular}{|c|c|c|}
    \hline
       &\textbf{\# of posts}  & \textbf{\# of sentences}  \\\hline \hline
       \textbf{Frankenstein}&2947&38432\\\hline
       \textbf{The Hobbit}&2897&37529\\\hline
       \textbf{Of Mice and Men}&2956&30205\\\hline
       \textbf{To Kill a Mockingbird}&2893&33000\\
       \hline
    \end{tabular}
    \caption{Data description and size.}
    \label{tab:my_label}
\end{table}
\par
Although our initial study corpus consisted of sixteen novels, we selected these four novels for detailed analysis on the basis of the broad disparity in their narrative structures, large variability in the number of characters, and a broad range of character relationships. For example, \textit{The Hobbit} can be characterized as a multi-episodic, linear narrative that takes place across many different settings in an elaborate fantasy world, and includes a large cast of both human and non-human characters, instantiating an elaborate version of a standard hero's journey plot. \textit{Of Mice and Men}, by way of contrast, is a short novella with a limited cast of characters that takes place in a highly localized, realistic setting, and represents a straightforward version of Vonnegut’s “From bad to worse” plot. \textit{Frankenstein}, although told partly in flashback, has a largely linear plot and a limited cast of characters, with a strong central figure and a relatively clear villain, although this is complicated by its use of nested narratives. Finally, \textit{To Kill a Mockingbird} has an overlapping set of complex characters with multiple subplots.
\par
For our ground truth narrative framework graphs, we relied on the online SparkNotes resource for each of the four chosen novels. SparkNotes is a corpus of freely available, professionally generated summaries of works of fiction, and provides us with a list of actants, as well as a chapter level plot summary. These fine-grained summaries allowed us to manually create an actant-relationship narrative framework graph for each novel. These ground truth graphs were coded independently by two experts in literature, and a third expert was used to adjudicate any inter-annotator disagreements.
\par
Reviewers who post to Goodreads have a variety of motivations for posting. The majority of reviewers use the site as part of a social network focused on reading, with the gender balance of active reviewers skewing slightly toward women \cite{thelwall2017goodreads}. There appear to be several categories of active reviewers on the Goodreads site, including students reviewing books as part of school assignments, members of book clubs, and people who aspire to become professional book reviewers. We make no discrimination as to classes of reviewers, but rather consider each review equally, as our goal is to understand the aggregate narrative model of a reviewed book.  At the same time, we recognize that reviews of a book are often conditioned by the pre-existing reviews of that same book, including reviews such as those found in SparkNotes, Cliff Notes, and other similar resources. In certain cases, we recognize that these reviews may be influenced by the filmed adaptations of the target novels or professionally written summaries. 

\section{Methodology}
\begin{figure}
    \centering
    \includegraphics[scale=0.3]{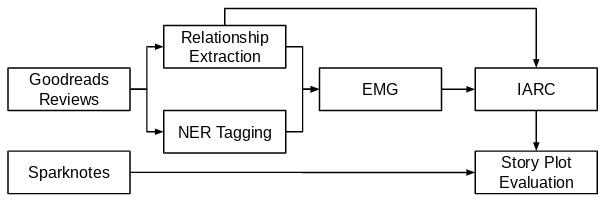}
    \caption{Pipeline to extract actant-relationship graphs. Our contributions introduce the Entity Grouping and the Inter-actant Relationship Clustering blocks}
    \label{fig:pipeline}
\end{figure}

Our methodology focuses on the underlying structure of the narrative framework that captures how a storytelling instance emerges via a collective negotiation process. Each post to a forum describes relationships among only a subset of actants (which are yet not known to our automated algorithms). 
To write a sentence, a reviewer first picks a context $C_i \in C$ and then samples an underlying context-dependent network $G_{C_i}(V_{C_i},E_{C_i})$ (to be estimated by the algorithm) by drawing a pair of actants $(A_k, A_j)$ according to a conditional actant recall distribution across all the actants, $p_{C_i}(A_j)$. A context could represent a particular situation in the plot. For example, when someone wants to recount the scene in \textit{Frankenstein} where Dr. Frankenstein creates the monster, then certain actants and relationships are described much more often than others. 

Following this, the reviewer draws a relationship for the pair $(A_k, A_j)$ from a distribution associated with the context-dependent edges: $D_{(E_{C_i},  (j,k))}({\mathcal {R}})$. The writer then composes the review according to these outcomes by choosing the proper words and syntax. In particular, the reviewer chooses noun phrases (as mentions of the actants $A_j$ and $A_k$) and the associated verb/relationship phrases (or other syntactical constructs) for the sampled relationship.

Recall that we have neither any knowledge of the underlying actants nor of the contexts that define different semantic relationships among them. After syntax-based relationship extractions from the reviews, we have multiple mentions/noun-phrases for the same actants, and multiple semantically equivalent relationship phrases to describe different contexts.  In order to accurately estimate the different contexts $C_i$, actant frequency distributions $p_{C_i}(A_j)$, and the relationships  $D_{(E_{C_i},  (j,k))}({\mathcal {R}})$, we must aggregate the different mentions of the same actant into a single group. In order to do that, we need to consider relationships: two mentions refer to the same actant only if the key relationships with other actants are semantically identical. Thus, the estimations of entity mention groups and relationships need to be done jointly.

The following subsections describe our approach to the estimation of the aggregate narrative network in the three steps of our pipeline presented in figure \ref{fig:pipeline}: (i) Syntax-Based Relationship Extraction, (ii) Entity Mention Grouping (EMG), and (iii) Inter-actant Relationship Clustering (IARC). The resulting graph constitutes an end-state ranked consensus model of all actants and relationships. The evaluation of our results focuses on the similarity of the ground truth and learned narrative graph based on a matching of actants and their contextual relationships. The frequency distributions of the actants, $p$, and relationships, $D$, can be estimated based on the counts of the occurrences of the associated groups of phrases. Currently, we use a threshold to decide whether an actant or a relationship is included in the consensus narrative graph. We leave a more detailed study of these frequency distributions and their relationship to reader consensus to ongoing and future work. These  probabilities encode the relative importance of the different actants and relationships in ways not captured by the thresholded network. For example, in The Hobbit,  the actant node ``Ring'' has only a single relationship edge   (i.e., ``Bilbo'' finds the ``Ring'') yet, due to the centrality of the ``Ring'' to the story, it has a frequency rank in the top ten among all noun phrases.  

\noindent \textbf{Syntax-Based Relationship Extraction}:
Each sentence in the text corpus is processed to extract specific patterns of syntax relationship tuples in the form of ($arg_1$, $rel$, $arg_2$) where arg1 and arg2 are noun phrases, and rel is a verb or other type of phrase. Our relation extraction combines dependency tree and Semantic Role Labeling (SRL) \cite{gildea2002automatic}\cite{manning2014stanford}. As opposed to limiting our extractions to agent-action-target triplets, we design a set of patterns (for example, Subject-Verb-Object (SVO) and Subject-Verb-Preposition (SVP)) to mine extractions from dependency trees using the NLTK package and various extensions. 
The patterns are based on extensions of Open Language Learning for Information Extraction (OLLIE) \cite{schmitz2012open} and ClauseIE \cite{del2013clausie}. Next, we form extractions from the SENNA Semantic Role Labeling (SRL) model. We combine dependency-based extraction techniques with SRL to increase the recall of our system. 
A list of all the syntax relationship patterns, their definitions, and related examples are provided in the GitHub link for our research.

Following these steps, we apply cleaning and de-duplication techniques to select unique and high precision extractions. Relationship tuples scraped from reviews only include those entity mentions that match or exceed a frequency lower bound ($\geq 50$).

\noindent \textbf{Entity Mention Grouping (EMG)}: As a semantically identifiable character in a book is expressed in reviews as diverse entity mentions, it is necessary to group these mentions and label them with the same character label. 

Let the frequently-occurring set of entity mentions be $M$ and let $R_{ik}$ be the relationships between entity mention $m_i$ and $m_k$, where $m_i$ is the Subject and $m_k$ be the Object. The set $R_{ki}$ then denotes the relationships when the roles are reserved. First, we note that if there is a relationship triplet $(\mbox{Subject}=m_i, \mbox{Verb}, \mbox{Object}=m_j)$  then clearly $m_i$ and $m_j$ are mentions of different actants and are not to be grouped together. In order to avoid any noise-induced exclusion of such a pairing, we  consider a pair $m_i, m_j$ as incompatible if  $|R_{ij}| + |R{ji}| \geq \gamma$. Based on our observation of the low frequency of noisy relationships, the  hyperparameter $\gamma$  is set to 3 in this paper. In the following we assume that for each mention $m_i$ we have removed all incompatible nodes $m_j$. 

Intuitively, two compatible mentions $m_i$ and $m_j$ correspond to the same actant if, for every other mention $m_k$, the relationships between the pair $(m_i, m_k)$ are semantically the same as the relationships between the pair $(m_j, m_k)$. In practice, different mentions of the same actant will share only a subset of the relationships when aggregated over all the extractions. In the following we provide an algorithm to quantify this intuitive idea that yields robust EMGs.

\begin{figure*}
    \centering
    \includegraphics[width=\textwidth, scale=0.2]{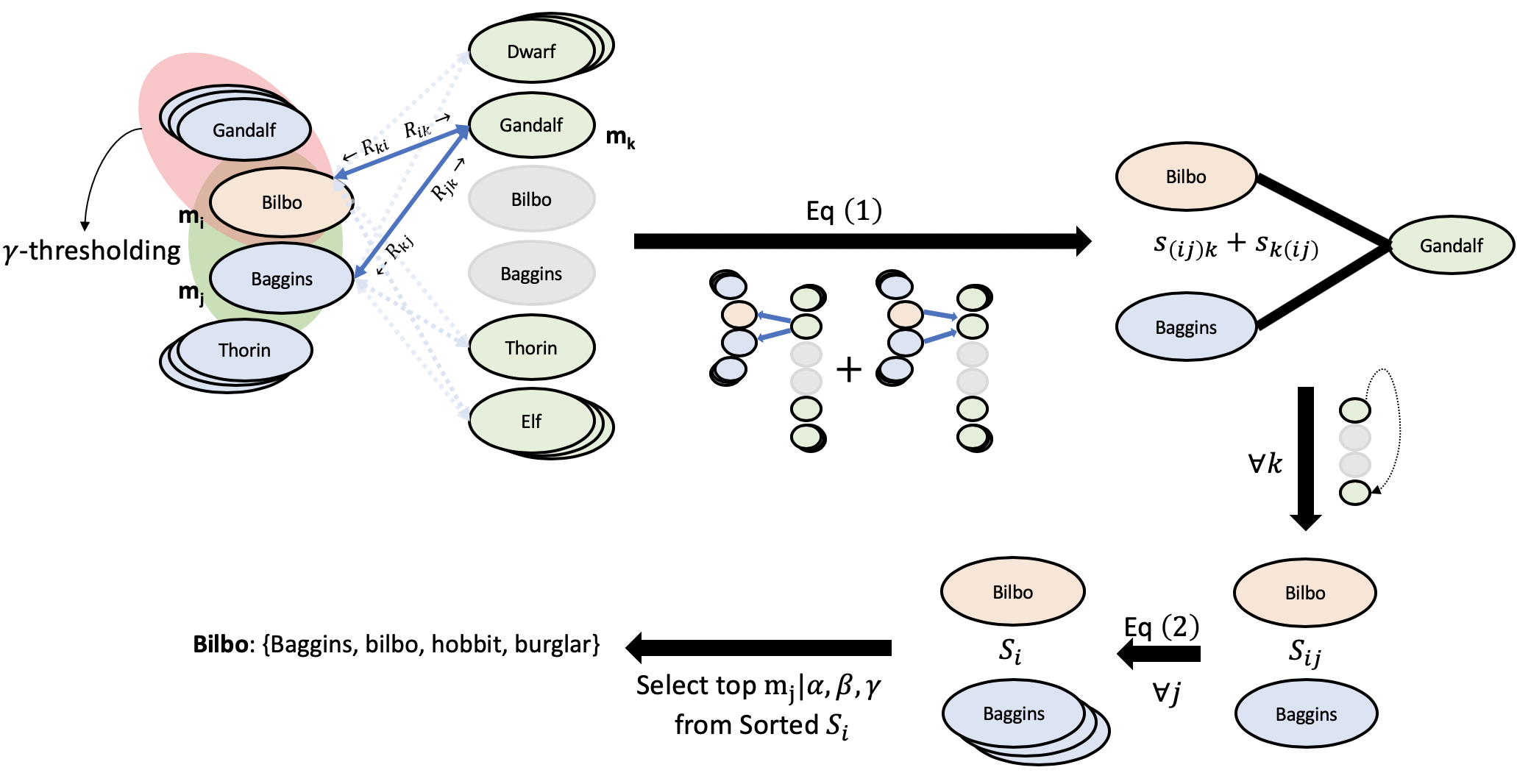}
    \caption{The pipeline of the EMG task shows the formation of the bipartite graph G with the computation of the Score Matrix $S$,  along with hyperparameters $\alpha, \beta, \gamma$}
    \label{fig:bi_graph}
\end{figure*}

Let $T_{ik} = H(R_{ik})$ describe the set of headwords in $R_{ik}$. Also let $G$ be the directed bipartite graph from the entity mentions $M$ to $M$ (see Fig.~\ref{fig:bi_graph}) with the edges representing the relationships between the entity mentions. We would like to find an Entity Mention Grouping (EMG) function $g:M \to [1,...,N]$, $N \leq |M|$, where (i) if  $g(m_i)=g(m_j) =k$ then entity mentions $(m_i, m_j)$ are grouped together to form the $k^{th}$ actant.  Moreover, (ii) we want the groups to be complete: that is, for two groups $g^{-1}(k_1)$ and $g^{-1}(k_2)$ (with $k_1\neq k_2$ and $k_1,k_2 \in [1,...,N]$), the entity mentions are semantically similar within each set and are semantically differentiated across the sets. To measure semantic similarity between $m_i$ and $m_j$, we consider the following measure involving another mention $m_k$:

\begin{equation} 
\begin{split}
s_{(ij)k} &= \Pr (T_{ik} | T_{jk}) + \Pr (T_{jk} | T_{ik})\ ,\\
\Pr (T_{ik} | T_{jk}) &= \frac{|H(R_{ik}) \cap H(R_{jk})|}{|H(R_{jk})|}.
\end{split}
\end{equation}

To understand why $s_{(ij)k}$ is an effective similarity measure, consider the following cases: 
(i) If $H(R_{ik})=H(R_{jk})$, implying that $m_i$ and $m_j$ share the exact relationships with $m_k$ and hence should be grouped together, then $s_{(ij)k}$ achieves the maximum value of 2, (ii) the $m_j$ mention of an actant occurs less frequently then $m_i$ and is reflected by $H(R_{ik}) \subset H(R_{jk})$, then $s_{(ij)k} \geq 1$. This captures the case where $m_j$ shares all its relationships with $m_i$ but not vice versa, (iii) $m_i$ and $m_j$ are indeed mentions of different actants, in which case  $|H(R_{ik}) \cap H(R_{jk})|$ is expected to be a lot smaller than both $|H(R_{ik})|$ and $|H(R_{jk})|$ and $s_{(ij)k} << 1$. 

To ensure that we compute  similarity when $m_k$ is the Subject, we define an analogous similarity score:
\begin{equation} 
\begin{split}
s_{k(ij)} &= \Pr (T_{ki} | T_{kj}) + \Pr (T_{kj} | T_{ki})\ ,\\
\Pr (T_{ki} | T_{kj}) &= \frac{|H(R_{ki}) \cap H(R_{kj})|}{|H(R_{kj})|}.
\end{split}
\end{equation}

Finally, the score matrix $S$ is computed where the score $S_{ij}$ between $m_i$ and $m_j$ aggregates the measure on all feasible $m_k \in M - \{m_i,m_j\}$ and provides a metric for similarity across all entity mentions:
\begin{equation}
S_{ij} = \sum_{m_k \in M - \{m_i,m_j\}} s_{(ij)k} + s_{k(ij)}.
\label{equ:similarity-metric}
\end{equation}

The grouping function $g$ is now constructed as follows: For every entity mention $m_i$, the scores in the vector $S_i$ are ranked in descending order. We next introduce two hyperparameters for each novel,  $\alpha, \beta \geq 0$, such that an entity mention $m_i$ is grouped with $m_j$ only if the  score $S_{ij}$ satisfies: $S_{ij} \geq \alpha$ {and}  $\frac{S_{i(j-1)}}{S_{ij}}   \geq \beta$ (for $j\geq 2$). 

We compute $\alpha$  from novel-specific distribution statistics. In particular, we compute the histogram of all non-zero $S_{ij}$ and compute  $\alpha$ as the $75$th percentile (i.e. $25\%$ of $S_{ij}$'s are $\geq \alpha$). For all considered books (except \textit{To Kill a Mockingbird} where $\alpha = 2.6$), $\alpha = 2.0$. The hyperparameter $\beta$ is set to $2$.

The  parameters $\alpha$ and $\beta$ are similar to those in works such as the Elbow K-Means method \cite{kmean}, in which $\beta$ correlates to inertia if the scores $S_i$ correlate to the distortion, and $\alpha$ provides a means of resolution if the elbow is unreliable (common in our model for rarer entity mentions).

The entity mention groups, once found, are labeled with the most frequent mention in the respective groups. Empirically, these automatically computed labels match the  ground truth entities as derived from SparkNotes.

\noindent \textbf{Inter-actant Relationship Clustering (IARC)}: The aggregated entity mentions captured in $g$ are fed back into the standard relationship extraction task. Then, the relationships aggregated between any pair of actants, represented by their respective entity mention groups (e.g.: $A_1 = g^{-1}(k_1)$ and $A_2 = g^{-1}(k_2)$) is computed as:
\begin{equation}
R_{A_1A_2} = \underset{p \in A_1, \, q \in A_2}{\cup} R_{pq}.
\end{equation}

$R_{A_1A_2}$ is a richer and potentially multi-modal set of relationships. This process enables a form of transfer learning, aiding relationship extractors in identifying connections at a higher semantic level of characters and not merely at the level of entity mentions. The associated relationship clusters are found using the cosine similarity measure in the BERT embedding space (Algorithm 1).

\begin{algorithm}[ht]
\SetAlgoLined

\KwResult{$C_{A_1A_2}$}
  $\hat{R}_{A_1A_2},C_{A_1A_2}$ = \{\}\;
  \For{$r \in R_{A_1A_2}$}{$
  \texttt{append }\, \texttt{BERT}(r) \texttt{ to }\, \hat{R}_{A_1A_2}$
  }
  $C_{A_1A_2}$ = \texttt{Elbow K-Means Method on } $\hat{R}_{A_1A_2}$ \\

 \caption{Inter-actant Relationship Clustering}
\end{algorithm}

$C_{A_1A_2}$ is the set of clusters of relationships that describe the multi-modality in $R_{A_1A_2}$. For each cluster  $C$ we compute its dispersion (using the cosine similarity measure), $\beta_C$. We retain only those clusters with  $\beta_C$ greater than a threshold (here, we set it to $0.8$) as a valid semantic relationship group.

\noindent \textbf{Evaluation: }
We compare these relationship clusters to the ground truth relationships between characters (e.g.: $J_{A_1A_2}$). We aim to find a mapping $h_{A_1A_2}:J_{A_1A_2} \rightarrow C_{A_1A_2}$. This process is described in Algorithm $2$, where $f_{cos}(a,b)$ is the function to compute the cosine similarity between $a,b$, and $\beta_C$ is the dispersion of a cluster $C$ using the cosine similarity measure. Thus, a ground truth relationship phrase is mapped to an automatically clustered semantic group only if its embedding is close enough to the centroid of the cluster. 
\begin{algorithm}[ht]
\SetAlgoLined

\KwResult{$h_{A_1A_2}$}
  \For{$C \in C_{A_1A_2}$}{
  
  \If{$\beta_C \geq 0.8$}{
  
  \If{$\underset{r \in C, \, j \in J_{A_1A_2}}{\max} f_{cos}(r,\texttt{BERT}(j)) \geq 0.8$}{
  $h_{A_1A_2}(j) = C$
  }
  }
  }

 \caption{Evaluation: Mapping Relationship Clusters to Ground Truth}
\end{algorithm}

Similar to the EMG task, the clusters are well differentiated, resulting in high-fidelity labels. Furthermore, Algorithm 2 seeks to approximate a maximum likelihood estimation problem, where $\mathcal{L}$ represents the cosine similarity $f_{cos}$ implemented with thresholds: 
\begin{equation}
h_{A_1A_2}(j)= \underset{C \in C_{A_1A_2}}{\textrm{argmax}} \,\, \mathcal{L}(C,j), \,\, \forall \, j \in J_{A_1A_2}.
\end{equation}

\section{Limitations}

Data can be noisy, particularly when social media posts, which are informal by nature, are the primary source. This informality creates noise in the relationship extraction phase. A missing punctuation mark, for example, can significantly change the dependency tree structure and lead to erroneous extractions of both the arguments and the relationship phrases. 

Other parts of our pipeline are equally sensitive to noise, including pronoun resolution and BERT embeddings. While pronoun resolution is needed to improve coverage (that is, to capture relationships amongst entity mention references when they are expressed in terms of pronouns), the process adds additional noise by occasionally resolving pronouns to the wrong entity mentions. Error from pronoun resolution is more noticeable in relation to rare words. For example, in the sentence, ``The example their single father Atticus sets for \textit{them} is one all parents wish we could parallel.", \textit{them} is mapped to the single character \textit{Dill}. \textit{Dill} is among the characters mentioned least frequently in reviews of \textit{To Kill a Mockingbird}. In such a scenario, the extracted relationships have a low fidelity because of the sparse sample space.  In addition, while the BERT embeddings that we use for this paper provide useful vectors in cosine-measured k-means clustering, the approach also suffers from sensitivity to noise. 

Using SparkNotes as a ground truth also raises some issues, as the summaries in these reader guides are less detailed than the novels that they summarize. Consequently, comparing our extractions to the limited relationships described in SparkNotes means that some of our discovered relationships, which may be in the novel but not in the SparkNotes summary, are improperly evaluated (i.e. the relationship exists in both the target novel and our extractions but is missing in SparkNotes). For example, while our extractions reveal that George cares for or loves Lennie in \textit{Of Mice and Men}, this relationship is missing from the SparkNotes summary. Similarly, certain actants or relationships that exist in the ground truth summaries may simply be absent from the reader review corpus, as is the case for certain Frankenstein actants such as M. Krempe. Our methods are not able to discover actants or relationships that do not appear in reader reviews--this elision of characters and relationships, however, may be indicative of interesting aspects of reader review practice.

\section{Results}
We first examine the syntactic method of establishing actant-actant relationships for clustering. In Table \ref{tab:appos}, the Appos and SVCop relationships suggest not only limiting sentence-level associations, but also semantically invariant associations mentioned explicitly in the reviews. While this syntactic approach may work in many situations, book reviewers often \textit{assume} a basic shared knowledge of the plot of a novel.  This assumption dissuades reviewers from explicitly writing out the relationships between actants. In addition, book reviews are not very descriptive in general, focusing more on specific plot points or a character's trajectory. This tendency in book reviews further weakens direct Appos and SVCop actant-relationship extraction.

{}

\begin{figure*}
    \centering
    \includegraphics[width=\textwidth, scale=0.5]{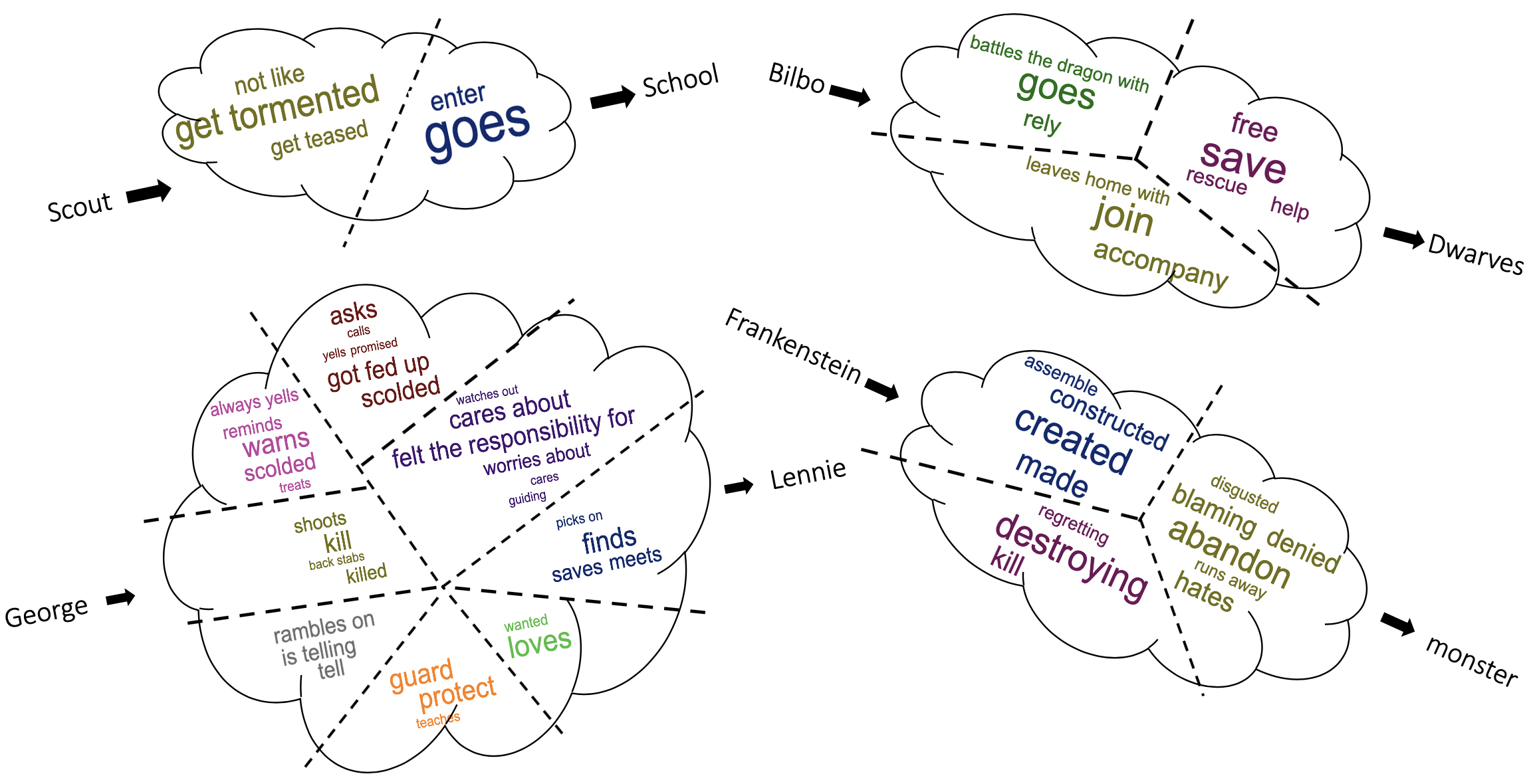}
    \caption{Directed and clustered relationships emergent after IARC between 2 actants per novel. In clockwise direction from top left: from Scout to School in \textit{To Kill a Mockingbird}, from Bilbo to Dwarves in \textit{The Hobbit}, from Frankenstein to Monster in \textit{Frankenstein} and from George to Lennie in \textit{Of Mice and Men}.}
    \label{fig:dir_rels}
\end{figure*}{}
\begin{table}[]
    \centering
    \begin{tabular}{|p{1.8cm}|p{6cm}|}
    \hline

     \textbf{Entity} & \textbf{Descriptors} \\
    \hline \hline
        & \textbf{The Hobbit}\\
    \hline
      Bilbo& (a, the, simple, clean) hobbit, a burglar, baggins, hero, protagonist\\
    \hline
    Smaug & (a, the, horrible, vicious) dragon\\
    \hline
    Gandalf & (a, the, wise) wizard \\
    \hline
    & \textbf{Frankenstein}\\
    \hline
     Frankenstein & (a, the, fantasy) book, (the, a) creator, (a, the) doctor \\
     \hline
     Monster & (his, a, the) creation\\
     \hline
    & \textbf{Of Mice and Men}\\
    \hline
     George & a small (man,-, in height), Lennie's (caretaker, best friend, father figure, protector)   \\
     \hline
     Lennie & (the, pitiful, unique, favorite) character, George's ( foil, best friend)\\

     \hline
     & \textbf{To Kill a Mockingbird}\\
    \hline
     Jem & (big, the older, strong) brother\\
     \hline
     Atticus & (the, loving, ordinary, her) father\\
     \hline
     Scout & (a, hotheaded, young, an interesting) Tomboy\\
     \hline
   
    \end{tabular}
    \caption{Examples for Appos and SVcop candidate descriptors for entity mentions across the four novels.}
    \label{tab:appos}
\end{table}

\begin{table}[]
\begin{minipage}[b]{1.0\linewidth}
\centering
\begin{tabular}{ | c | c | }
    \hline
    \textbf{Entity Mention} & \textbf{Ranked Similarity Scores} \\
    ($m_i$) & \textbf{for other Mentions ($m_j$)} \\
    & \textbf{($S_{ij}$'s, see Eq.~\ref{equ:similarity-metric})}  \\ \hline \hline
    Bilbo &  baggins,42.14\\
     & hobbit,14.47 \\
     & burglar,3.80  \\\hline
    Burglar & bilbo,3.80\\
    & dwarves,2.79\\ \hline
    Wizard &gandalf,22.49\\
    & gandolf,7.00\\
    &grey,5.34\\
    &thorin,3.32\\\hline
    Hobbit&bilbo,14.47\\
    &baggins,6.06\\\hline
    
   \end{tabular}
    \caption{Given two entity mentions $(m_i,m_j)$, the similarity score $S_{ij}$ (see Eq.~\ref{equ:similarity-metric}) measures the semantic ``fungibility'' of the mentions (i.e., whether both  mentions are used interchangeably to refer to the same actant). The table shows several popular  entity mentions ($m_i$'s) and the similarity scores of other candidate mentions, $m_j$'s, in \textit{The Hobbit}. Clearly, the mentions [Bilbo, baggins, Hobbit, Burglar] form a clique representing the same actant, \textit{Bilbo Baggins}. One can also see the emergence of another EMG [Wizard, Gandalf, Gandolf, Grey] for the actant \textit{The wizard}. }
    \label{table:ranked candidates}
\end{minipage}

\begin{minipage}[b]{1.0\linewidth}
\centering

\includegraphics[scale=0.4]{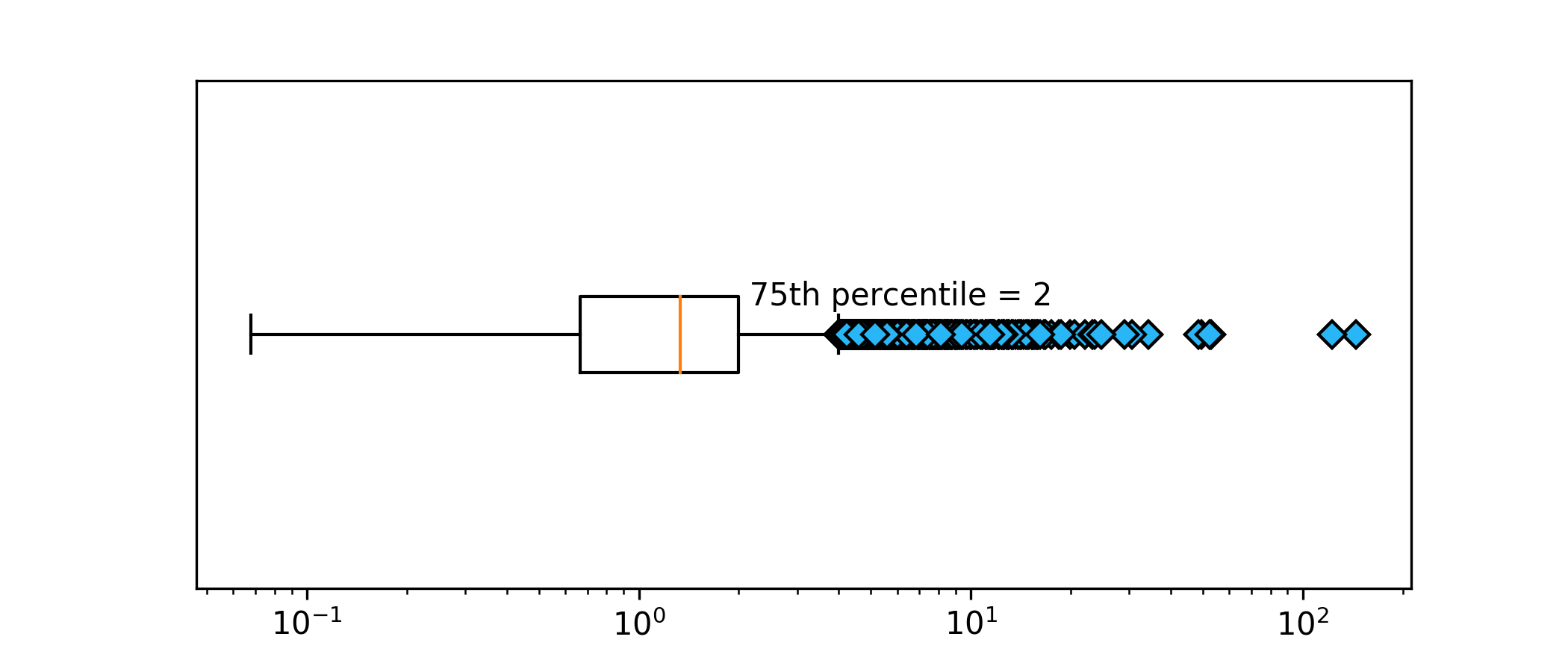}
\captionof{figure}{A Box plot of the similarity scores, $S_{ij}$'s (see Eq.~\ref{equ:similarity-metric}), for all entity mention pairs $(m_i,m_j)$  in \textit{The Hobbit}. For any entity mention, $m_i$, its Entity Mention group (EMG) is first pruned to contain $m_j$'s with scores, $S_{ij}\geq \alpha$, where $\alpha$ is the $75^{th}$ percentile of the score distribution. From the plot we find $\alpha=2$. This EMG  is further pruned by first sorting the list by their scores, and then ensuring that the ratio of any two successive scores is bounded below, i.e., $\frac{S_{i(j-1)}}{S_{ij}}   \geq \beta$ (for $j\geq 2$). We found that $\beta =2$ provided a good cutoff.     }
\label{fig:hist scores}
\end{minipage}
\end{table}

We applied our EMG algorithm to obtain the actants as documented   in Table \ref{tab:entity groups}. Table~\ref{table:ranked candidates} and Fig.~\ref{fig:hist scores} provide example statistics obtained during the execution of the EMG algorithm.  Each actant, once formed,  aggregates relationships that the individual entity mentions imply.
The clustering of relationships aggregated under the now-formed entity mention groups yield higher granularity and confidence in the IARC task, as semantic connections between entity mentions reinforce the relationships from one actant to another. This effect is observed across the four books as shown in Fig. \ref{fig:dir_rels}. The relative size of words in the figure correlate to their frequency in the aggregated relationships between the entity mention groups. 

\begin{figure}
    \centering
    \includegraphics[scale=0.3]{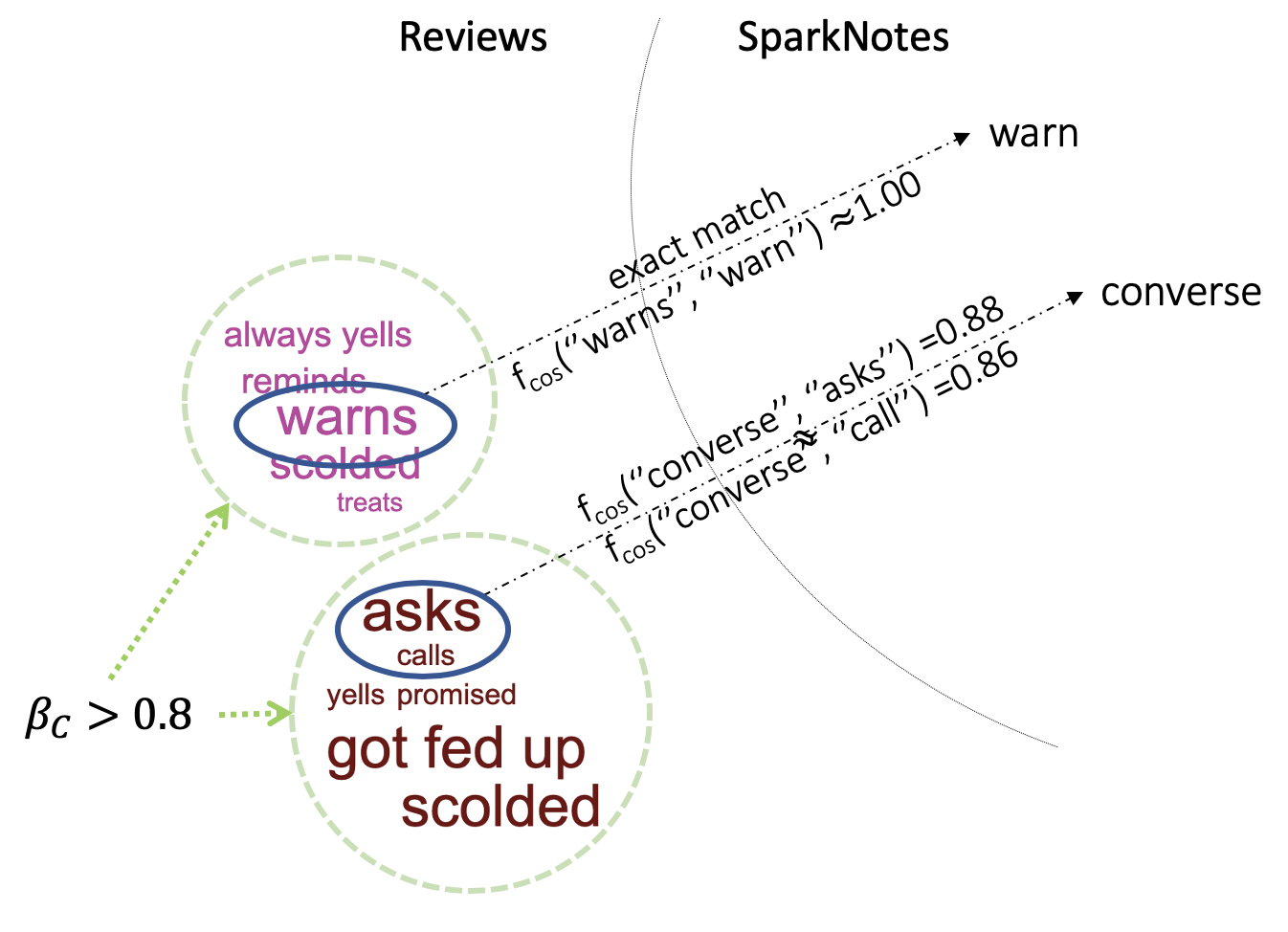}
    \caption{Evaluation phase: matching 2 clusters of relationships in Of Mice and Men, from George to Lennie, to ground truth labels, in accordance to Algorithm 2. $\beta_c$ determines the set of edges.}
    \label{fig:cosine}
\end{figure}

The task of mapping relationship clusters to particular ground truth labels is shown for the ``converse'' and ``warn'' clusters from George to Lennie in \textit{Of Mice and Men} (Figure \ref{fig:cosine}).  The rich clusters, in comparison to the ground truth labels from SparkNotes suggests recall as a good measure of performance for our pipeline. A summary of our results for all four books including recall is presented in Table \ref{tab:table_2}.

In general, the relationships between actants reveal a high degree of consistency with the ground truth graph. The largest divergences consist of missed relationships rather than the identification of non-existent relationships, although these occur occasionally. This latter group of relationships is often the attribution of a relationship, such as the killing of Smaug (the dragon in \textit{The Hobbit}), to an important character such as Bilbo Baggins. In other words, many readers \textit{incorrectly believe} that Bilbo killed Smaug. Another small set of spurious relationships, including one that suggests that Jem killed Bob Ewell in \textit{To Kill a Mockingbird}, are caused by reader confusion, “what-if” scenarios or, more commonly, incorrect pronoun resolution and aggregation. Apart from the relatively infrequent misattribution of relationships, the reduction in relationships aligns with the corresponding reduction in the number of actants connected to the central component of the story graph. 

\begin{figure}
    \centering
    \includegraphics[scale=0.15]{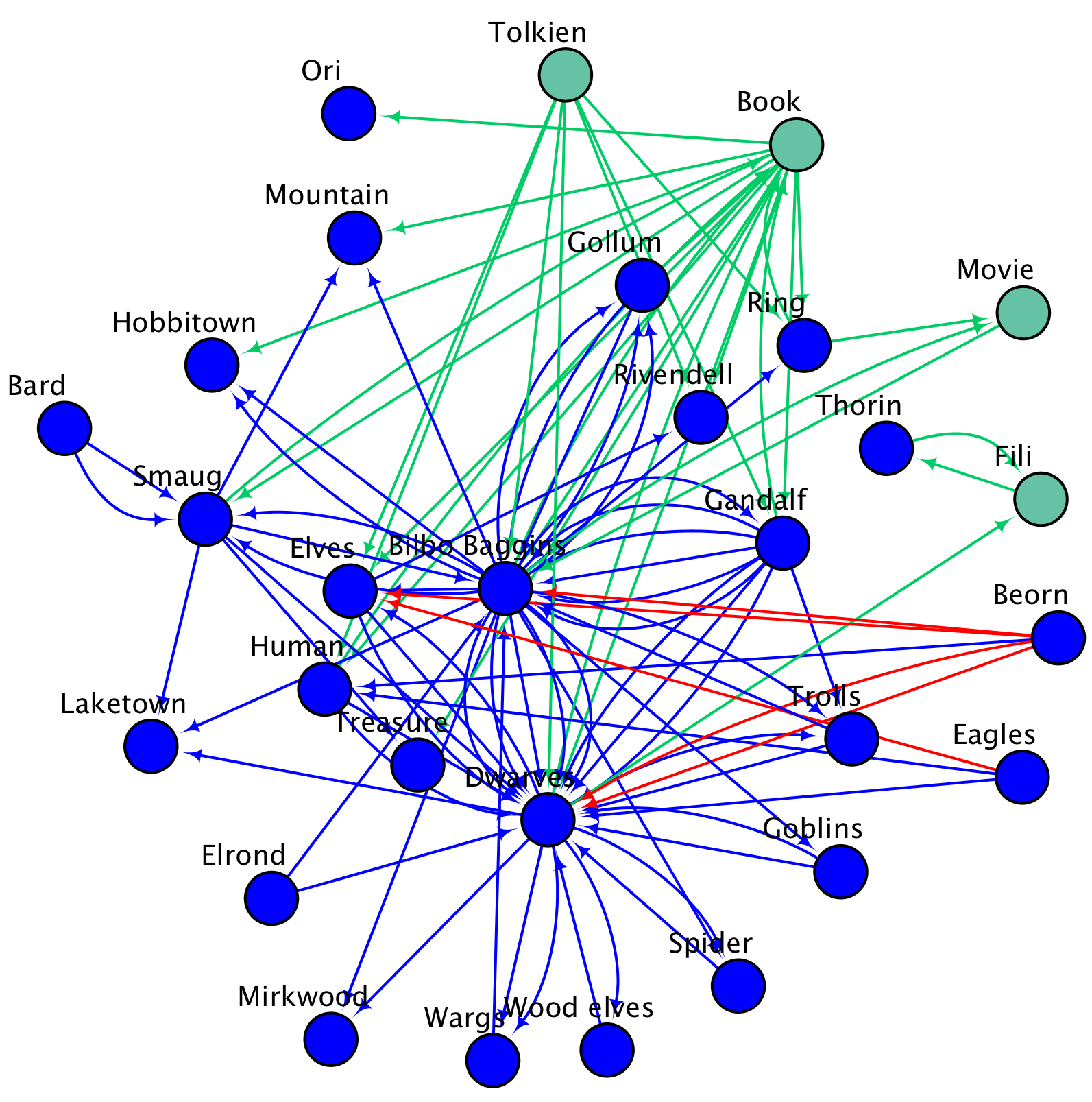}
    \caption{Narrative Framework graph of \textit{The Hobbit}. Green nodes are extracted entities not part of the ground truth, red edges are ground truth edges which were not detected by the algorithm, blue edges are detected ground truth edges.}
    \label{fig:story_1}
\end{figure}
\begin{figure}
    \centering
    \includegraphics[scale=0.15]{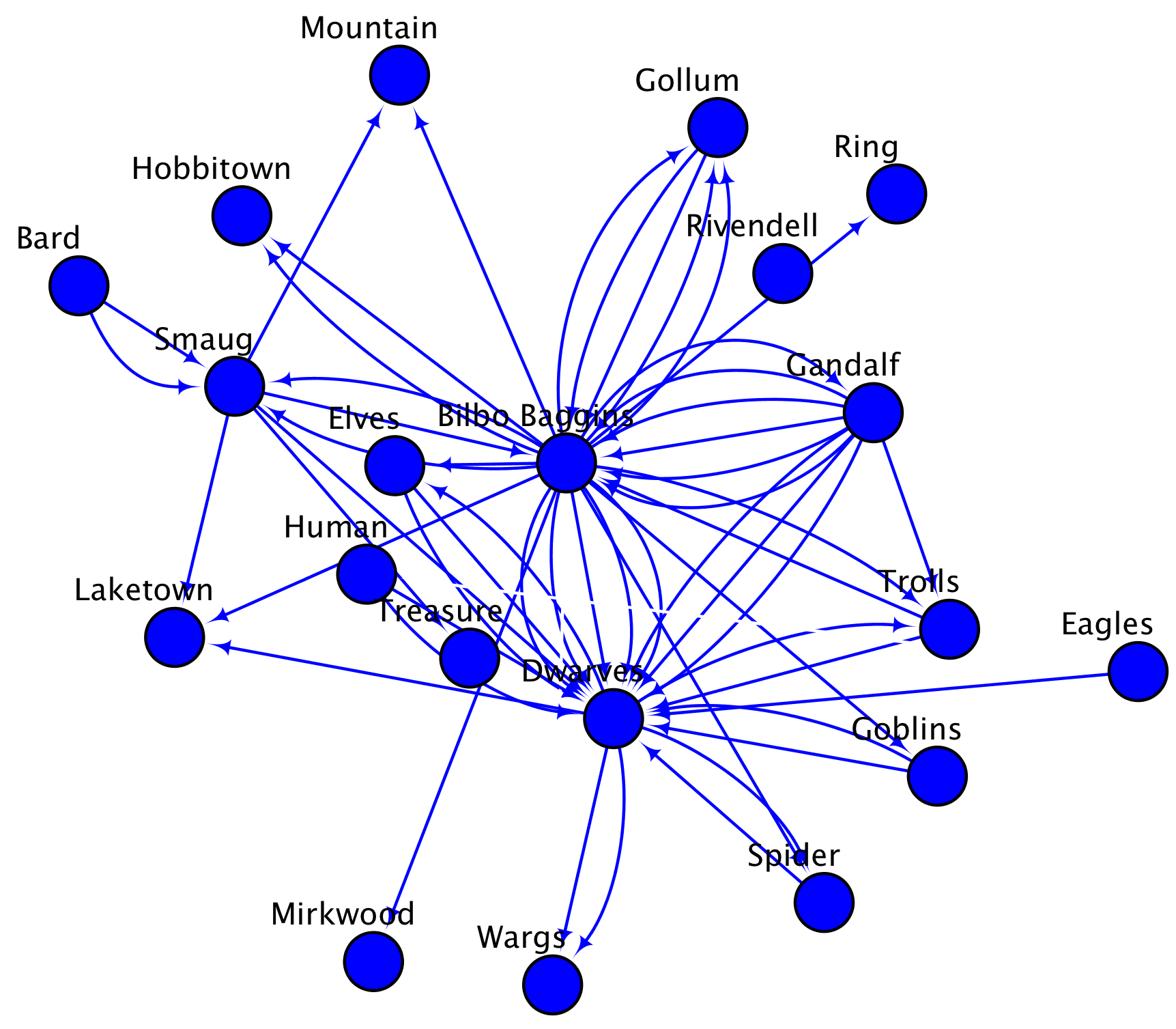}
    \caption{Narrative Framework graph of \textit{The Hobbit} \textit{after} thresholding on the frequency of relationship. Blue edges have at least 5 relationship instances.}
    \label{fig:story_2}
\end{figure}
Figure \ref{fig:story_1} depicts the narrative framework graph for \textit{The Hobbit} with blue nodes representing ground truth actants or meta-actants. We also show four examples of resolved actants or meta-actants (colored green) not found in the ground truth: \textbf{Tolkien}:[tolkein, author], \textbf{novel}:[book, fantasy, story, novel], \textbf{Fili}:[fili]  and \textbf{Film}:[film, movie, scene]. 
Blue edges represent relationships in the ground truth found by using our methods (frequency threshold $\geq 5$), while red edges represent undetected ground truth relationships. Green edges connecting to green nodes (frequency threshold $\geq 10$) are edges that cannot be verified; we include them to indicate the richness of the extracted graph as opposed to the ground truth. Figure \ref{fig:story_2} shows a graph similar to Figure \ref{fig:story_1} after the deletion of low frequency edges ($\leq 5$), and represents the core structure of the narrative covered in the reviews conditioned on the SparkNotes ground truth.

There are shared structural properties (disregarding the specific relationships they encode) that can be used to automatically distinguish between actual characters in the novels and the various meta-actants. For example, the meta-actant \textbf{Tolkien}  (the green node at the top center of Figure \ref{fig:story_1}) has only outgoing edges, indicating that Tolkien appears only as the subject in any inferred relationship triplet. This lack of incoming  edges is a significant feature of meta-actants: An important character in a novel usually has  bi-directional relationships with  other characters. An author of the novel, on the other hand, usually ``acts'' on the characters; hence the corresponding node is directionally isolated. The incoming edges for the meta-actant ``Book'' are all attributable to phrases such as " character XNZ is  portrayed \textit{in the} book/novel''. A simple filtering of these preposition-induced relationships directionally isolates the meta-actant ``Book.'' Further structural explorations of the derived networks, such as measures of centrality and importance of different characters, are part of our ongoing work.

\begin{center}
\begin{table*}[t]
\large
    \begin{tabular}{|p{3cm}|p{14cm}|}
    \hline
    \textbf{Book Name} & \textbf{Entity Mention Groups}  \\
    \hline \hline
     {\textbf{Of Mice and Men}} & \textbf{Lennie}:[Lennie, lenny], 
    \textbf{George}:[george, milton], 
    \textbf{Curley's Wife}:[curley's wife, tart, wife], 
    \textbf{Aunt Clara}:[aunt clara, aunt, clara],
    \textbf{men}:[workers, men],
    \textbf{ranch}:[ranch, farm],
    \textbf{soft things}:[soft things, soft, things],
    \textbf{mental disability}:[mental disability, mental, disability]\\
    \hline
    {\textbf{The Hobbit}} & \textbf{Bilbo}:[bilbo, baggins, burglar, hobbit],
    \textbf{Rivendell}:[rivendell, middleearth],
    \textbf{Gandalf}:[gandalf, wizard, gandolf, grey],
    \textbf{dwarf}: [dwarf, dwarves],
    \textbf{Thorin}: [thorin, company],
    \textbf{trolls}:[trolls, orcs],
    \textbf{elf}:[elf, elves],
    \textbf{Hobbitown}:[hobbitown, shire, hobbiton],
    \textbf{man}: [human, man, lakemen],
    \textbf{dragon}:[dragon, smaug]\\
    \hline
    {\textbf{Frankenstein}} & 
    \textbf{monster}:[monster, creature, adam], 
    \textbf{Frankenstein}:[frankenstein, victor, doctor, creator],
    \textbf{Mary Shelley}: [mary, shelley, author, mary shelley],
    \textbf{Elizabeth}:[elizabeth, wife],
    \textbf{Walton}:[walton, robert],
    \textbf{Henry}:[henry, clerval],
    \textbf{Justine}:[justine, moritz],
    \textbf{Caroline}:[caroline, beaufort]
    \\
    \hline
    {\textbf{To Kill a Mockingbird}} & 
    \textbf{Scout}:[scout, sister], 
    \textbf{Atticus}:[atticus, dad, father, finch], 
    \textbf{Jem}:[jem, brother],
    \textbf{Harper Lee}: [lee, harper lee, author, harper],
    \textbf{Tom}: [tom, robinson, negro, mockingbird, africanamerican],
    \textbf{Bob}:[bob, ewell],
    \textbf{Boo}: [boo, arthur, arthur radley, boo radley],
    \textbf{Mayella}: [mayella, daughter],
    \textbf{aunt}: [aunt, alexandra],
    \textbf{Maycomb}: [maycomb, alabama, town],
    \textbf{Heck}:[heck, tate],
    \textbf{Cunningham}:[cunningham, walter]\\
    \hline
    \end{tabular}
        \caption{Final actants after EMG per book. Each actant group is labeled with the most frequent mention in the group. Empirically, these automatically computed labels match the  ground truth entities as derived from SparkNotes. }
    \label{tab:entity groups}
\end{table*}
\end{center}
\begin{table*}[]
\resizebox{\textwidth}{!}{
\begin{tabular}{|l|l|l|l|l|}
\hline 
 & \textbf{Of Mice and} & \textbf{The Hobbit} & \textbf{Frankenstein} & \textbf{To Kill a} \\
 & \textbf{Men} & & & \textbf{Mockingbird} \\
 \hline \hline
\textbf{Recall (\%)}                 & \textbf{88.33} (83.33)                 &\textbf{82.61} (59.42)           & \textbf{69.04} (66.66)               & \textbf{90.16} (68.85)               \\ \hline
\textbf{Edge detection rate (\%)}   & \textbf{98.33} (96.66)                & \textbf{92.75} (69.56)           & \textbf{73.80} (73.80)                & \textbf{93.44} (77.04)              \\ \hline
\textbf{Average Number of Relationships} & \textbf{246.55} (209.15)               & \textbf{139.34} (14.03)           & \textbf{20.33} (13.38)              & \textbf{72.09} (27.34)               \\ \hline
\textbf{Median Number of Relationships}  &  \textbf{54} (48)                & \textbf{43} (3)           & \textbf{7} (7)                 & \textbf{36} (6)                \\ \hline
\end{tabular}}
\caption{Performance on character relationship extraction with IARC after (\underline{in bold}) and before (\underline{within parentheses}) EMG. In the ``before'', scenario an actant group consisted of only the mention used in the ground truth. Thus for actant ``Bilbo'' only the mention ``Bilbo'' was used to compute its relationship. Post EMG, the mentions in the group \textbf{Bilbo}:[bilbo, baggins,burglar,hobbit] were aggregated to compute the actant Bilbo's relationships.}
\label{tab:table_2}
\end{table*}

\section{Discussion}
The results support the idea that readers, when summarizing a novel, tend to reduce the scope of the story and to focus on the most memorable aspects of the plot, here modeled as inter-actant relationships. In the reviews we studied, people converge on a set of main actants and relationships that map well to a core set of actants and relationships in the ground truth summaries, suggesting that people are relatively adept at summarizing even complex novels. As part of their summaries, however, people tend to simplify. This simplification may be related to cognitive limits on the number of real-world relationships that a person can keep in mind. 

Since reviews tend to be short, when compared to the length of the work summarized, it is not surprising that people reduce both the number of actants, particularly in works with very large casts of characters such as \textit{The Hobbit}, and the relationships between those actants. The inter-actant relationships are also simplified in the reader reviews. Readers can simplify complex plots, such as that in \textit{To Kill a Mockingbird}, into relatively straight forward stories of conflict, strategies to address that conflict, and the result of the use of those strategies. The reduction of plot complexity may also be influenced by the abstraction of the novel in other media. For certain books, such as \textit{The Hobbit}, recent films have been highly successful, and it is quite possible that movie watching has had some impact on reader reviews. The same may apply to the other books in this study given, for example, the numerous references to the actor Gregory Peck in the reviews of \textit{To Kill a Mockingbird}. Although we have not done so here, it may be interesting to compare reader reviews of filmatized novels to the summary story graphs for those films.

\section{Conclusion}
The approach we describe here is widely applicable to other crowd-sourced review sites such as Rotten Tomatoes and Metacritic (for films) and LibraryThing and Love Reading (for literature) that, much like Goodreads, allow viewers or readers to present their own reviews of fiction, be it literature or film. An intriguing aspect of many of these sites is the propensity of reviewers to provide ``plot summaries'' as opposed to critical engagements with more sophisticated thematic analysis. While this plot-based approach to reviewing works of fiction may drive literary scholars to the brink of insanity, it does allow us to consider questions regarding the popular engagement with literature and other forms of artistic production. In future work, we expect to include actant-relationship sequencing so that we derive automatically a reader consensus model of plot, represented as a dynamic narrative framework graph. Given the responses that people post, we can use the scale of these sites to derive insight into how people (or groups of people) not only read but also remember. Turning the process around, it may be possible to develop a dynamically updated crowd-sourced summary of a novel or film--as more people write reviews, the consensus summary would update, capturing the emphasis on actants, relationships, and events that commentators add. Such a system could act as a cultural response barometer since what people remember, and what they forget (or choose to leave out), can be telling indicators of popular engagement with art.

\nocite{tangherlini2016mommy}
\bibliography{sample}
\bibliographystyle{plainnat}

\end{document}